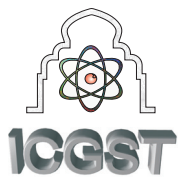

www.icgst.com

# Visual Simplified Characters' Emotion Emulator Implementing OCC Model

Ana Lilia Laureano-Cruces[1,2,3], Laura Hernández-Domínguez[2], Martha Mora-Torres[2], Juan-Manuel Torres-Moreno[3] Jaime Enrique Cabrera-López[2]

[1] *Departamento de Sistemas, Universidad Autónoma Metropolitana – Azcapotzalco,*
*San Pablo 180, CP.02200. México, DF.*

[2] *Posgrado en Ciencia e Ingeniería de la Computación, Universidad Nacional Autónoma de México,*
*Edif. Anexo IIMAS, 3er piso, Ciudad Universitaria, Coyoacán, México, D.F. C.P. 04510*
clc@correo.azc.uam.mx (lili94@exalumno.unam.mx)
laudobla@gmail.com, kabhun@yahoo.com.mx

[3] *Laboratoire Informatique d'Avignon / Université d'Avignon et des Pays de Vaucluse*
*BP 91228, 84911 Avignon Cedex 9, France*
juan-manuel.torres@univ-avignon.fr

**Abstract**
In this paper, we present a visual emulator of the emotions seen in characters in stories. This system is based on a simplified view of the cognitive structure of emotions proposed by Ortony, Clore and Collins (OCC Model). The goal of this paper is to provide a visual platform that allows us to observe changes in the characters' different emotions, and the intricate interrelationships between: 1) each character's emotions, 2) their affective relationships and actions, 3) The events that take place in the development of a plot, and 4) the objects of desire that make up the emotional map of any story. This tool was tested on stories with a contrasting variety of emotional and affective environments: Othello, Twilight, and Harry Potter, behaving sensibly and in keeping with the atmosphere in which the characters were immersed.

## 1. Introduction

Throughout history, Artificial Intelligence has sought to create machines that emulate intelligent behavior; in particular, it has attempted to repeat in the computer common behaviors of humans. The latter pursuit has resulted in efforts to create systems that can *think* and act like humans, following equanimous logical rules and norms oriented toward the fulfillment of well-defined objectives and goals. However, people are complex beings, whose actions do not depend only on measurable and provable logical chains of reasoning, but rather our emotions play a fundamental part in the decision-making process, and consequently in our actions and reactions. The latter are seen as powerful heuristic mechanisms that permit a decision-making process when there are numerous options [13, 17]. Examples of application include those related to the teaching-learning process [9, 10, 15].

Consequently, it is to be supposed that to create machines that think and act like humans, we will need to contemplate the factors that make us act as humans, such as awareness and emotions. The inclusion of awareness in the development of intelligent agents is a very recent field that is only beginning to be studied and analyzed from a computational perspective, and one of the first approaches to that objective was made by [3], where *emotions* are one of the characteristics used to accomplish it. However, the inclusion of emotions in this area has been studied previously, and more extensively, by [16] in their theoretical design of the cognitive structure of emotions, the investigation forming the basis of our discussion in this paper and which we will refer to henceforth as the *OCC Model (Ortoney, Clore, Collins)*.

The article is organized as follows: *section two* offers an explanation of OCC Model and mentions research related to that approach. *Section three* describes the Visual Simplified Characters' Emotion Emulator (EVE) system modeled based on simplified OCC theory. In section four we present and analyze scenarios from three perspectives: events, objects, and actions of agents, exemplified with the Othello characters [18]. We explain the logic behind the system, meaning how quantitative reactions are formed to: events, objects, and agents. Finally, in the conclusions we discuss the implications for improving the software system; given that, in our view, the OCC Model provides an invaluable resource for modeling emotions from a cognitive perspective.

## 2. Overview of the OCC Model

The OCC Model proposes that we approach emotions in a delimited and formal context to establish a contained and well-defined structure that considers a finite number of possible situations that can trigger emotions [16]. To achieve this, the authors rule out aspects such as physiology, behavior, and linguistic expressions which are linked to emotions, and if considered can influence the decision-making process. Instead of considering these aspects, the *OCC Model* addresses emotions as specific types of cognitive processes based on personal and interpersonal descriptions of each situation, and therefore the results depend largely on the particular way each person can conceptualize each situation, and may take varying forms depending on the personal idiosyncrasies of the designer of the structure.

The *OCC Model* was not created with the intention of making a computer actually *feel* emotions, but rather of making it, *interpret the ways emotions can be included and evaluated by humans. Last has an effect in humans' action*. It can also be used to imitate a behavior that adapts to a given emotion of the user. With this in mind it is possible to take into account the events of a dynamic environment, and create interfaces that can interact with humans more realistically.

### 2.1 Principal Aspects of the World According to the OCC Model

The OCC Model postulates that emotions have a certain intensity or valence and that any reaction or emotion corresponds to an intensity with direction from the perspective of the standpoint of the person experiencing it. This model, takes into account the world under three central aspects to consider when dealing with emotions and the factors that trigger them: events (related to goals), agents (related to actions), and objects (related to the ability to attract).

*Events*: are things that happen without necessarily requiring the intervention of an agent, for example when it starts to rain or a relation dies.
The main feature in assessing an agent's emotional reaction to an event is its *desirability*. Thus, the death of a loved one is a highly undesirable event, whereas the recovery of a loved one after a prolonged illness is highly desirable.

*Agents*: are the main actors that will have some form of participation or causality in the emotional structure. Agents are not limited exclusively to persons, but in some cases there may also be *inanimate objects or abstractions considered agents*, as long as they can be considered actuators in their specific context.

Agents are characterized by their ability to generate actions that change the course of events in a story and its *emotional map*. The actions each agent can take have a value of plausibility that makes it possible to determine the effect they will have when performed by a particular agent. For example, helping an old woman cross the street is a highly plausible event, while stealing candy from a child is a censurable event. Obviously, our assessment of an action's degree of plausibility will depend on each person's moral values and interpretations.

*Objects*: are manifestations that are immersed in an environment and do not have an action or active participation in that environment, but which, nonetheless, have a certain level of attraction that transforms them into objects of desire or repulsion that can have a powerful influence on the emotional atmosphere. Objects are not necessarily physical manifestations; they can also be *goals, desires, or even persons or ideas*. Figure 1, shows a synthesis of the relevant aspects of the OCC Model [15, 16]

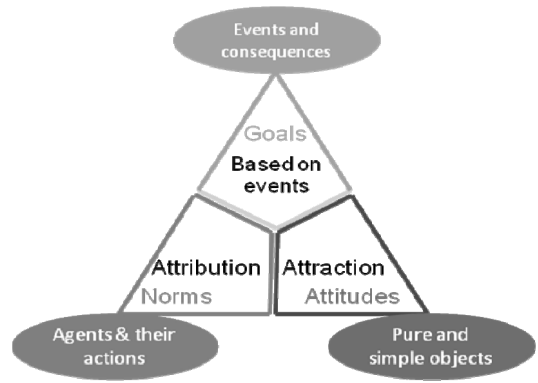

Figure 1. The OCC model theory

### 2.2 Related Work

The *OCC Model* provides the primary foundation for our discussion in this paper. However, the system was constantly tested throughout its development and improvement for the classic story of Othello [18], based on the mental models proposed for the play by [6] and an approach based on cognitive and affective motivational structures [1, 2, 7, 8, 9, 10, 15].

The OOC Model has been utilizing successfully to generate behavior, which is conjoined with the external aspects of emotions. The above with and to generate an action consistent with the feelings who make the decision, [1, 2, 4, 5, 7, 8, 10, 12]

## 3. Description of EVE

Visual Simplified Characters' Emotion Emulator (EVE) was programmed in the Adobe Flash

development environment due to its excellent suitability for creating visual interfaces.

This system was designed so that the user *can create emotional and affective environments in real time*, designing, with the system's help, the characters that take part in the story, their interpersonal affective relationships, their possible actions, events, and the events and objects that will form part of the emotional architecture of each story

### 3.1 Design of Agents

For this system the agents considered are restricted to human characters from stories, unlike agents in the OCC Model which *may also be inanimate objects*. EVE was designed to emulate the emotional behavior of *two, three, or four characters* from a particular story. The limit on the maximum number of characters is due to an esthetic aspect of system output, as a larger number of characters on screen produce an overcrowded and unclear visual interface. The real-time design of interpersonal relationships among agents by users also influenced this factor. The user must describe each of the relationships between characters, and a larger number of agents would represent a profusion of relationships that could overwhelm the user. A pending task involves simplifying the graphic display.

The system allows the user to characterize some components of the agents, including: name, hairstyle and color, and eyeglasses, in order to establish stronger identification between the agent and the character he represents (Figure 2).

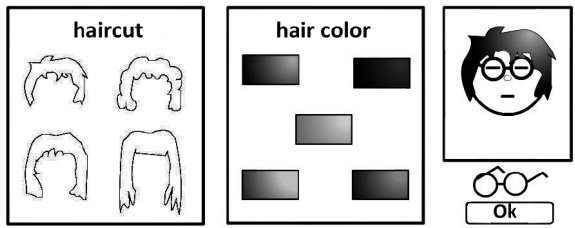

Figure 2. Interface to design agent's physical appearance

In addition to the characters' physical appearance, the user can also define affective relationships between the agents (Figure 3). The range of values corresponding to affection ranges from -5 (total hatred), passing through 0 (indifference), to +5 (unconditional love/friendship). The system does not prompt the user to define an agent's level of affection toward himself given that, to reduce the exponentially of options, *the system operates on the premise that all agents love themselves unconditionally*.

The assessment of interpersonal affections between agents is very important given that they constitute the basis for obtaining emotions of good will in the future, such as: happiness for someone's good fortune, or ill will, such as: happiness in another's misfortune.

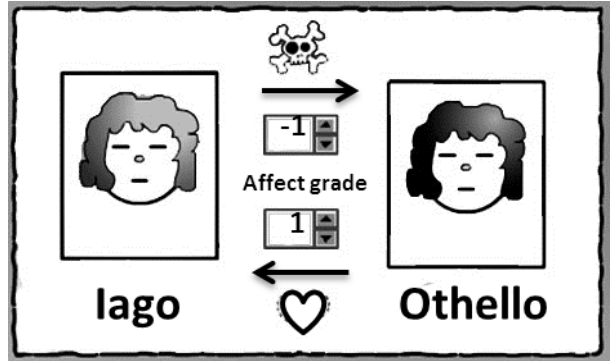

Figure 3. Design of affective relationships between agents

### 3.2 Design of Events

EVE sees events as things that happen in the course of a story, but are not direct consequences of the agents' actions. Events are an environmental factor external to the characters which, notwithstanding, can affect them.

The system shows the user an interface where she can assign a name or title to an event, as well as its level of desirability (Figure 4), with a range varying between -5 (undesirable) and +5 (very desirable). It is important to point out that, for this system, the level of desirability of an event is the same for everyone, *which is not true in real life, as each person interprets an event from a different perspective which implies a different level of desirability*. Again, this restriction is imposed because varying levels of desirability of events for agents would result in a wide variety of assignments that could overwhelm users when they design the environment.

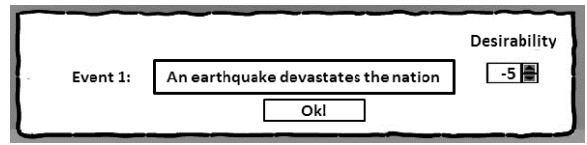

Figure 4. Event design interface

### 3.3 Design of Objects

As in the OCC Model, for the purposes of this system, objects are elements of the world that may be material (money, a house) or immaterial (goals, desires, abstractions). In the interface, objects are designed in a manner similar to that used to design events: they are assigned a name and a level of appeal or desirability ranging from -5 (a repulsive object) and +5 (a fascinating object).

### 3.4 Design of Actions

In the OCC Model actions are part of the agent behavior. The characterization of actions in EVE has been separated from the characterization of agents. Actions are defined as acts that are available for any of the agents to perform. Their characterization requires an interface identical to those used for objects and events, except that in this case the degree of plausibility of an event is assigned as a value in a range from -5 (a highly censurable event) to +5 (a praiseworthy event).

Design of actions by the user concludes the section on design of the environment and begins the stage of design of emotions. In any case, the system gives the user the option *to go back at any time* and change the values assigned to the different elements of the environment.

### 3.5 Emotional Map

Having designed: 1) agents and their relationships, 2) events, 3) objects, and 4) actions that will come into play when testing the system, we can observe a general map of the actual emotions of each of the characters that have been designed, the relationships between the characters, and the chances for realization of events, objects, and actions. The resulting map is shown in Figure 5.

In this map (Figure 5) there is a large icon pertaining to each of the central characters, accompanied by a group of smaller icons showing the faces of the other characters. These smaller icons show the other characters' mindset in relation to the central character. The right side of the map shows the *events, objects, or actions that were previously defined by the user* (defined in other screens), as well as their respective desirability, appeal, and plausibility (represented by a whole number in the range of -5 to 5). Also, a character can be selected to be the center: 1) of an event, 2) acquire an object, 3) perform an action, or 4) be affected by an action. This is achieved by clicking on the character's large icon.

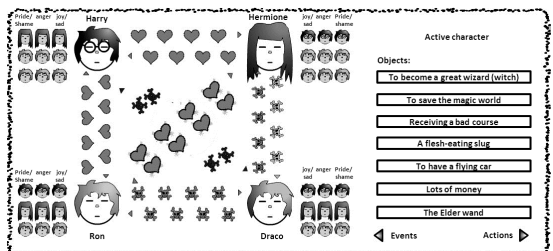

Figure 5. Overall emotional map for the story of Harry Potter

### 3.6 Emotions Considered in EVE

There is a wide variety of emotions that can be experienced by a person, or a character in a story. To restrict emotional relationships, only three emotions were chosen: *happiness, anger, and pride;* with levels of intensity ranging from -5 to +5.

*Happiness*: is, in fact, a two emotions representation: happiness and sadness. The range considered for this emotion is from -5 (distress), passing through 0 (neutrality) to +5 (euphoria). The faces depicting each of these emotions are shown in Figure 6.

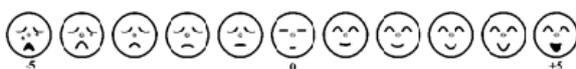

Figure 6. Representative faces for the range of values for the emotion *sadness-happiness.*

*Anger*: is an emotion with a range of values varying from 0 (calm) to 5 (rage). There are no negative values for this emotion and its graphic representation is shown in Figure 7.

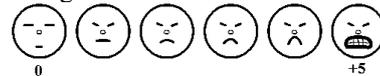

Figure 7. Representative faces for the range of values for the emotion *anger.*

*Pride*: like happiness, pride is represented by two emotions: pride and shame. The range of values considered for this emotion varies from -5 (shame), passing through 0 (neutrality) to +5 (pride). The faces depicting each of these emotions are shown in Figure 8.

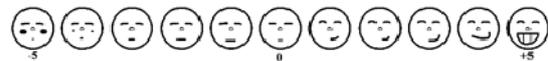

Figure 8. Representative faces for the range of values for the emotions *shame-pride.*

It is important to mention that, because this is a simulator of emotions in stories, the characters *feel* an emotion, but it does not disappear unless an event alters it; in other words, *a historical record is kept of each character's emotional status*, and therefore if a character is sad and there is an event triggering happiness at a given time, that character will only feel happy to the extent that his previous sadness permits. Stated differently, an emotion felt previously can alter the intensity with which a given character experiences a new emotion, thereby presenting *conflicting feelings*.

### 3.7 Affective Relationships

Affective relationships among characters also have a visual representation, as an aid to the user's understanding. Figure 9 shows each of the possible affective links there may be between the characters in a story.

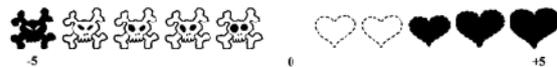

Figure 9. Visualization of affective relationships between characters.

## 4. Possible Scenarios

To better understand the work performed and the results obtained, below we show a series of *possible scenarios*, which allow us to visualize some of the emotional reactions of characters in the story of Othello [18] in response to certain events. The last is based in the emotive macro-structure designed in [6]. These scenarios were designed only to exemplify possible events, but practically any story can be designed differently *based on the reader's (user's) interpretation*. The visual design of the characters is shown in Figure 10.

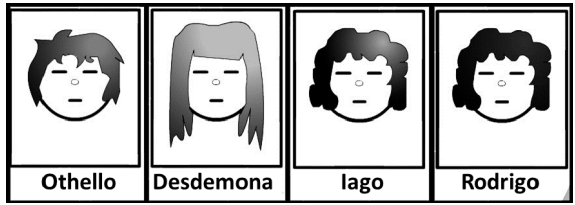

Figure 10.Visual design of characters in Othello

On the other hand, the affective relationships between the characters in Othello are shown in Table 1.

Table1. Affective relationships between characters in Othello.

|  | Othello | Desdemona | Iago | Rodrigo |
|---|---|---|---|---|
| Othello |  |  |  |  |
| Desdemona |  |  |  |  |
| Iago |  |  |  |  |
| Rodrigo |  |  |  |  |

The characters' emotions evolve as the story progresses. Accordingly, these examples assume that each character *always starts from a neutral value* in their emotions, as a reference point prior to the change triggered by the *event, object, or action* that has appeared.

### 4.1 Emotions Triggered by Events
Suppose that we have the following situation: Desdemona suffers a misfortune (an event with as desirability value of -5), such as *her father gets angry with her and opposes her marrying* Othello. Then, after that event, her emotional environment is expressed in the system as shown in Figure 11.

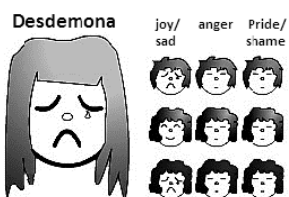

Figure 11.Desdemona's emotional map after she suffers a misfortune.

As the image shows, Desdemona's main emotional state is sadness. We can also see that Othello and Rodrigo love her unconditionally, and feel sadness for her in the same degree, given that, because they love her, they have good will toward her. However, we can also see that Iago feels happy at Desdemona's misfortune. His happiness can be explained by the fact that Iago does not love Desdemona and has ill will toward her.

### 4.2 Emotions Triggered by Objects
Imagine now a different scenario, in which an *object of desire* is in play, as occurs in the emotional map obtained when Rodrigo receives the rank of lieutenant (Figure 12).

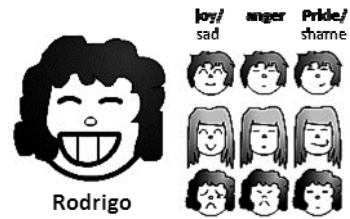

Figure 12.Rodrigo's emotional map after obtaining a highly appealing object

Achieving a goal for Rodrigo means that his primary emotional state shown in the system is pride. Also, because Othello and Desdemona have a degree of admiration for Rodrigo, they respond proportionally to their affection toward: happy and proud because of his achievement. Iago, who hates Rodrigo, feels angry (frustrated) and sad because of the benefit Rodrigo has received.

### 4.3 Emotions Triggered by the Characters' Actions (Agents)
The characters' actions trigger an emotional reaction *somewhat more complex than those triggered by objects and events*. This is because actions involve *two characters* (the character performing the action, and the character that action affects). Evidently, in a story the character performing an action and the character affected may be the same. This is exemplified in Figure 13, which shows the result obtained after Iago performs highly censurable action toward Othello: *betrayal*.

Analyzing the emotions depicted in Figure 13, we can see that Othello's primary emotional state is sadness at having been the object of a *censurable action*. Also, we can see that Desdemona and Rodrigo also feel a certain degree of sadness at Othello's misfortune, and in turn feel anger toward Iago for having harmed a loved one. Othello also feels anger toward Iago, but also, because he likes him, *feels shame* because of the censurable nature of the action performed by someone he admires.

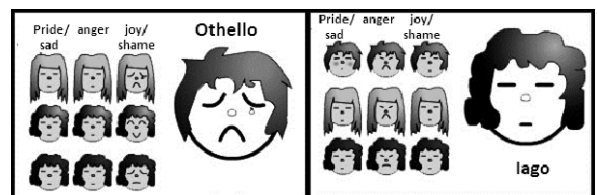

Figure 13.Emotional reactions after Iago's betrayal of Othello

### 4.4 The Logic behind the System
The characters' reactions in the course of a story are kept and intensify as it progresses. To reflect this, a new emotional value is added to each character's emotional state, which corresponds to the new event that has occurred in the story. This is accomplished by merely adding the interpretation of the resulting emotion to the current emotion of the character in question.

A visual interpretation of this is provided by a simple example: suppose that Harry Potter has just lost one of his loved ones. This leaves him completely devastated and, although Gryffindor will win the house cup, something that has made him feel extremely happy on several occasions, his emotional state does not reach extreme happiness, given that in the *conflicting feelings of losing a loved one and winning the house cup*, the former outweighs the latter. This is shown in Figure 14.

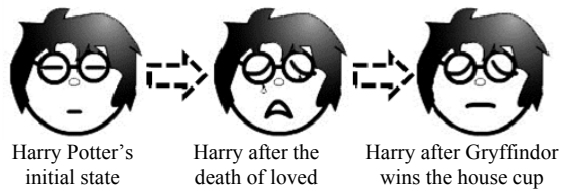

| Harry Potter's initial state | Harry after the death of loved one | Harry after Gryffindor wins the house cup |

Figure 14. Emotional progress of character Harry Potter through a sequence of events

The interpretation of the resulting emotion *in response to an event, object, or action* is based on *matrices of quantitative reactions with interpretative values*. These matrices have been filled in by the development team based on their personal experiences and their own interpretation of emotional reactions to the elements in question.

To better understand this, the Table 2 presents the matrix of *characters' quantitative reactions to events*. For the time being these matrices are equal for all characters. In this matrix the *rows* represent the emotional level of the character reacting to the character, to whom the event occurs. The *columns* represent the desirability of the event. Thus, the quadrant of the matrix shaded *light gray* represents the section of *ill will*, where a misfortune produces the maximum increase in happiness for the person who loathes the person affected by that event. In contrast, the quadrant shaded *dark gray* represents feelings of *good will*, meaning that a character is cheered by another character's good fortune.

Table 2. Matrix of quantitative reactions to events. *x axis*: emotional value. *y axis*: desirability of the event.

| x\y | -5 | -4 | -3 | -2 | -1 | 0 | 1 | 2 | 3 | 4 | 5 |
|---|---|---|---|---|---|---|---|---|---|---|---|
| -5 | 5 | 4 | 3 | 2 | 1 | 0 | -1 | -2 | -3 | -4 | -5 |
| -4 | 4 | 3 | 2 | 1 | 1 | 0 | -1 | -1 | -2 | -3 | -4 |
| -3 | 3 | 2 | 2 | 1 | 1 | 0 | -1 | -1 | -2 | -2 | -3 |
| -2 | 2 | 2 | 1 | 1 | 1 | 0 | -1 | -1 | -1 | -2 | -2 |
| -1 | 1 | 1 | 1 | 1 | 1 | 0 | -1 | -1 | -1 | -1 | -1 |
| 0 | 0 | 0 | 0 | 0 | 0 | 0 | 0 | 0 | 0 | 0 | 0 |
| 1 | -1 | -1 | -1 | -1 | -1 | 0 | 1 | 1 | 1 | 1 | 1 |
| 2 | -2 | -2 | -1 | -1 | -1 | 0 | 1 | 1 | 1 | 2 | 2 |
| 3 | -3 | -2 | -2 | -1 | -1 | 0 | 1 | 1 | 2 | 2 | 3 |
| 4 | -4 | -3 | -2 | -1 | -1 | 0 | 1 | 1 | 2 | 3 | 4 |
| 5 | -5 | -4 | -3 | -2 | -1 | 0 | 1 | 2 | 3 | 4 | 5 |

We can also see that a character that has no emotional value toward another character; equally has no reaction to *fortunate or unfortunate events* affecting that character. Similarly, an event that is considered neutral (neither fortunate nor unfortunate) produces no reaction in the characters.

In total, *eight matrices of quantitative reactions* were generated: a) One to account for characters' feelings toward events (happiness; shown in Table 2), b) three for reactions to objects (shame-pride, sadness-happiness, and anger), c) one matrix for happiness in the person affected by actions, and d) three for emotional reactions toward the perpetrator of an action (shame-pride, anger, and sadness-happiness).

## 5. Conclusions and Future Work

This paper lays the foundation for *analysis of characters' emotional reactions* in stories, and may also help to improve interpretations of users' emotional states or in developing interfaces in which stories take place, such as automatic generation of stories and development of videogames.

The OCC cognitive theory of emotions [16], is a methodology used to evaluate the possible emotion from the cognitive point of view. Because this methodology is clear, precise and free of context has been used frequently in the synthesis of emotions by computer, [1, 2, 4, 5, 7, 8, 10, 12]

This paper is a simplification of the *OCC Model*, but can be enriched with aspects that allow for modification of characters' initial emotional states, through interpersonal relationships, in order to consider in stories variants of intensity of feelings that often change over the course of a story. For example, if we assume that two characters hate one another (from the outset), the story's evolution could allow them to become great friends due to actions and events that take place in the plot (or vice-versa). Another possible extension would be to use NLP techniques to extract the emotions of textual documents in order to enrich the model. We think, statistical algorithms for automatically identification of opinions from unstructured text documents [20] could be adapted and used for this task.

The system shows the like examples the relationships of the protagonists of three stories: Othello, Twilight, and Harry Potter [14,18,19]. A pending task involves simplifying the graphic display.

The game can be found at the following sites:
- http://describe.iingen.unam.mx/%7Elhernandezd/Interfaces/emulador.html
- http://ce.azc.uam.mx/profesores/clc/ (software section).

## 6. Acknowledgements

This work is part of the Soft Computing and Applications research within the Emotions research section, funded by the Universidad Autónoma Metropolitana, also recognize the enthusiasm of Laura Hernández-Dominguéz (CVU: 328836); in software development. Martha Mora-Torres (CVU:

167259) and Jaime Cabrera-López (CVU: 268998); all students at the Postgrado en Ciencia e Ingeniería de la Computación at Universidad Nacional Autónoma de México and supported by Consejo Nacional de Ciencia y Tecnología-MEXICO).

**Biographies**

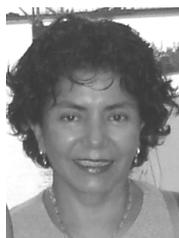 Ana Lilia Laureano-Cruces obtained a bachelor's degree in Civil Engineering, a Master's degree in Computer Sciences, and a Doctoral degree in Science at UNAM. She has been a full-time professor at Universidad Autónoma Metropolitana since 1987. She has more than 50 publications in the field of artificial intelligence. Between 1998 and 2000, she was a guest researcher at the Departamento de Sistemas del Instituto de Automática Industrial del Consejo Superior de Investigaciones Científicas de España in Madrid. Her principal research areas are: Interactive Learning Environments, Multi-Agent Systems, Intelligent Systems applied to Education, Expert Systems and Affective Computing.

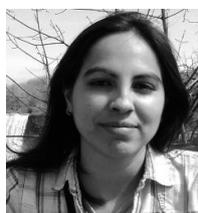 Laura E. Hernández-Domínguez is a computer engineer from Universidad Tecnológica de la Mixteca, Mexico. She is currently studying a Master in Computer Engineering and Sciences at UNAM, specialising herself on Computational Linguistics. She took part on the Student Design Competition of the Computer-Human Interaction 2008 conference, winning the first place. She also obtained a scholarship to attend the Grace Hopper Conference for Women in Computing 2007 and has attended to a variety of Mexican and international conferences on Information Technologies.

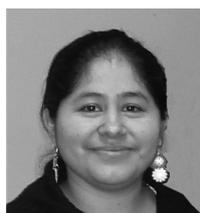 Martha Mora-Torres obtained a bachelor's degree in Electronic Engineering at Universidad Autónoma Metropolitana, and a Master's degree in Computer Sciences, at UNAM. She is currently a Doctoral student at UNAM. Her research interests are: Multi-Agent Systems, Intelligent Systems applied to Education, Expert Systems and Affective Computing.

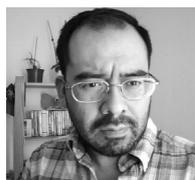 Juan-Manuel Torres-Moreno obtained a bachelor's degree in Electronic Engineering at Univer-sidad Autónoma Metropolitana, a Master's degree in Computer Sciences at INPG Grenoble (Fran-ce) and a PhD in Computer Science at INPG Grenoble. He has been a full-time professor at Université d'Avignon et des Pays de Vaucluse since 1993. He has more than 80 publications in the fields of Natural Processing Language and Machine Learning. He is currently the head of NLP Group at Laboratoire Informatique d'Avignon/UAPV (France). His research interests are: NLP, Intelligent Systems, Expert Systems and Affective Computing.

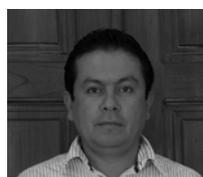 Jaime-Enrique Cabrera-López obtained a bachelor's degree in Electronic Engineering at Universidad Autónoma Metropolitana. He is currently studying a Master Sc. degree in Computer Engineering and Sciences at UNAM. Her research interests are: Multi-Agent Systems and Affective Computing.